\documentclass{iopconfser}
\usepackage{url}
\usepackage{graphicx}
\usepackage{lscape}
\begin{document}

\title{Summary of the Unusual Activity Recognition Challenge for Developmental Disability Support}

\author{Christina Garcia$^{1}$, Nhat Tan Le$^{2}$, Taihei Fujioka$^{1}$, Umang Dobhal$^{1}$, Milyun Ni'ma Shoumi$^{1}$, Thanh Nha Nguyen$^{2}$ and Sozo Inoue$^{1}$}

\affil{$^1$Graduate School of Life Science and Systems Engineering, Kyushu Institute of Technology, Wakamatsu, Japan}
\affil{$^2$Department of Biomedical Engineering, Ho Chi Minh City University of Technology, Ho Chi Minh, Vietnam}

\email{
garcia.christina-alvarez199@mail.kyutech.jp, alvarez7.christina@gmail.com}

\begin{abstract}
This paper presents an overview of the “Recognize the Unseen: Unusual Behavior Recognition from Pose Data” Challenge, hosted at ISAS 2025. The challenge aims to address the critical need for automated recognition of unusual behaviors in facilities for individuals with developmental disabilities using non-invasive pose estimation data. Participating teams were tasked with distinguishing between normal (e.g., sitting, eating) and unusual (e.g., head banging, attacking others) activities based on skeleton keypoints extracted from video recordings of simulated scenarios. The dataset reflects real-world imbalance and temporal irregularities in behavior, and the evaluation adopted a Leave-One-Subject-Out (LOSO) strategy to ensure subject-agnostic generalization. The challenge attracted broad participation from 40 teams applying diverse approaches ranging from classical machine learning to deep learning architectures. Submissions were assessed primarily using macro-averaged F1 scores to account for class imbalance. The results highlight the difficulty of modeling rare, abrupt actions in noisy, low-dimensional data, and emphasize the importance of capturing both temporal and contextual nuances in behavior modeling. Insights from this challenge may contribute to future developments in socially responsible AI applications for healthcare and behavior monitoring.
\end{abstract}

\section{Introduction}
Individuals with developmental disabilities often have unusual or atypical behaviors that require continuous monitoring and timely intervention to ensure their safety. These behaviors, which include stereotypical movements such as head banging, throwing objects, striking others, and biting fingers, present significant challenges to caregivers and support facilities \cite{fujioka1}. Early recognition of such unusual activities is essential to provide appropriate interventions and prevent potential harm to the individual and their caregivers. Unusual activity recognition in the context of developmental disability involves the detection of behaviors that deviate from typical patterns, including self-stimulating behaviors, self-injury actions, and other atypical movements \cite{vyas}. These behaviors are often repetitive and can be early indicators for interventions needed in developmental disability support \cite{gao}.

Facilities that support individuals with developmental disabilities often struggle to monitor and identify unusual behaviors due to limited staffing and the irregular nature of such actions \cite{fujioka1}. Traditional observation-based methods are time-consuming, error-prone, and insufficient for real-time intervention, creating a critical gap in care provision \cite{prakash}. The importance of unusual activity recognition extends beyond safety concerns. Continuous monitoring can provide valuable insights into behavior patterns, help optimize treatment strategies, and support research into developmental conditions \cite{stenum}. Early detection of unusual behavior can also facilitate prompt medical intervention, potentially preventing the escalation of distressing situations and improving the overall quality of life for individuals with developmental disorders. This has motivated the development of automated Human Activity Recognition (HAR) systems that can reliably detect abnormal behavior, especially in sensitive care environments \cite{fujioka1}.

Current HAR datasets face significant limitations when applied to the context of developmental disabilities specially the scarcity of annotated datasets. Most existing datasets focus on common everyday activities and do not adequately represent the wide range of unusual behaviors seen in individuals with developmental disabilities \cite{washington, morey}. In addition, annotation of unusual behaviors requires specialized medical expertise, making the process costly and time-consuming \cite{zhang}. The difficulty in obtaining large-scale, high-quality datasets for unusual activities creates a significant barrier in developing robust recognition systems. These limitations highlight the need for specialized data sets that accurately represent the unique characteristics of unusual activities in the context of developmental disabilities.

To address such limitations, this first International Symposium on Applied Science (ISAS) 2025 Challenge introduces a benchmark task that focuses on distinguishing typical daily activities from unusual behaviors using pose estimation data. The dataset consists of key points extracted from video recordings of healthy participants simulating eight activities: four normal activities and four unusual activities. This dataset demonstrates real-world complexities such as class imbalance (normal behaviors occur more frequently than unusual behaviors), irregularity and unpredictability of abnormal actions, temporal variability (abrupt versus sustained activity patterns), and subject variability (differences in body size, posture, and movement execution). These activities were specifically designed in consultation with staff in actual treatment facilities and reflect common behaviors that require monitoring and intervention in developmental disability support settings. The main objectives of this challenge include:
\begin{itemize}
\item Develop a model that can accurately classify normal and unusual activity from anonymized keypoint data.
\item Design a model that generalizes across individuals using a Leave-One-Subject-Out (LOSO) evaluation strategy.
\end{itemize}


Recent advances in HAR provide strong foundations for this challenge. Early video-based methods used 3D CNNs (C3D) to jointly model appearance and motion \cite{tran}, followed by two-stream networks combining RGB and optical flow \cite{simonyan}. Temporal Segment Networks (TSN) introduced sparse sampling for efficient long-sequence recognition \cite{wang}, while more recent transformer-based models, such as Video Swin Transformer \cite{liu}, achieve state-of-the-art results by capturing long-range temporal dependencies with hierarchical self-attention.

Although video-based methods are powerful, they also raise privacy concerns and face difficulties in healthcare applications where raw video recording of individuals with developmental disabilities may not be feasible. This has motivated increased interest in pose-based HAR, where 2D/3D skeleton key points derived from pose estimation serve as the primary input. The pose-based approach preserves anonymity while retaining important structural and temporal motion cues. Spatial-Temporal Graph Convolutional Network (ST-GCN) \cite{yan} pioneered the use of graph structures to model human joints and their temporal evolution, demonstrating robust performance on skeletal datasets such as NTU RGB+D. Extensions such as AS-GCN introduced adaptive graph learning to dynamically adjust joint connections \cite{shi}. More recently, PoseC3D \cite{duan} re-frames skeletal sequences as 3D heatmap volumes, achieving better robustness to noise and better generalization across datasets.

Pose-based HAR has advanced significantly, yet applying these methods to developmental disability support introduces unique challenges. The ISAS 2025 Challenge addresses this gap by providing a benchmark dataset focused on rare, safety-critical behaviors such as head banging, striking, and throwing objects rather than routine activities. Using pose-only data preserves privacy while pushing participants to tackle class imbalance, subject variability, and temporal uncertainty. The Leave-One-Subject-Out (LOSO) protocol further emphasizes generalization to unseen individuals, a critical requirement for real-world care environments. Overall, the challenge shifts evaluation from controlled benchmarks to socially impactful scenarios, aiming to advance not just accuracy but also robustness, scalability, and deployability of HAR systems in sensitive healthcare contexts.

The rest of this paper is structured as follows. Section 2 reviews related work on human activity recognition and pose-based approaches. Section 3 describes the challenge setup, including the dataset, goals, and evaluation protocol. Section 4 presents team performances on given tasks, and Section 5 summarizes the applied methods. Section 6 concludes with key insights and future directions.


\section{Related Work}
Recent studies in human activity recognition (HAR) highlight critical challenges such as irregular activities, data imbalance, temporal variation, and subject variability, which this challenge directly addresses.


\subsection{Irregular and Unpredictable Activities in HAR}
The recognition of irregular and unpredictable human activities remains a substantial challenge in HAR, particularly within care settings and populations exhibiting atypical behaviors, such as individuals with developmental disabilities. Routine, repetitive movements are frequently the focus of traditional activity recognition systems. Atypical or aberrant behaviours, on the other hand, such as aggressiveness, self-harming behaviour, or sudden changes in posture, are usually short-lived, intermittent, and situation-specific. It is challenging to properly record and characterise these behaviours since they rarely exhibit predictable motion patterns and can happen suddenly. Real-world behavioural datasets frequently exhibit a notable imbalance between normal and abnormal actions, which is further worsened by the lack of annotated abnormal event data \cite{fujioka1}.


Current HAR systems use vision or sensor-based deep learning methods to recognize activities \cite{fujioka1, c}. In large-scale contexts, models such as CNN-RF hybrids or YOLO detect anomalies from video \cite{fujioka1}, while skeletal pose estimation is preferred in smaller-scale settings for its privacy-preserving nature. Sequential models like LSTMs and Spatio-Temporal GCNs effectively capture temporal dynamics in posture or sensor data \cite{fujioka1, c}.


Unusual activity recognition remains constrained by data scarcity, as rare and unpredictable behaviors lead to imbalanced datasets and weak generalization of supervised models \cite{fujioka1, c}. Many approaches depend on handcrafted rules or image/video features that are noise-sensitive and overlook higher-level behavioral context \cite{c}. To address these issues, recent work explores temporal optimization, skeleton-based features, few-shot learning, adaptive windowing, and large pre-trained models to better capture irregular activities and improve generalization \cite{fujioka1}.

\subsection{Data Imbalance between Normal versus Unusual Activities}

One of the central challenges in HAR, especially in healthcare contexts, is the imbalance between frequent normal activities and rare unusual behaviors. In care facilities, most actions are routine (e.g., sitting, walking), while unusual behaviors such as head banging or self-injury are scarce, making models biased toward majority classes and reducing sensitivity to safety-critical events \cite{fujioka1}.  

Current solutions include controlled data collection, oversampling, SMOTE, GANs, and more recently, LLM-generated synthetic data to augment minority classes \cite{f}. However, these methods often face issues such as oversimplification, redundancy, unrealistic distributions, or high computational costs \cite{e}. Increasing sample volume alone is insufficient; adaptive learning, few-shot techniques, and domain-aware augmentation are needed to improve minority detection while preserving authenticity.  

Despite advances, models still struggle with temporal and subject variability, where abrupt unusual events, long normal actions, and individual-specific motion patterns challenge generalization \cite{fujioka1}. In this challenge, participants confront datasets dominated by normal actions but containing rare, transient anomalies, requiring models to handle imbalance, temporal irregularity, and subject-specific patterns. Promising strategies include adaptive or multi-scale windowing, minority-class augmentation, and invariant representation learning \cite{h}.

\subsection{Temporal and Subject Variability}

Temporal variability in HAR arises from differences in speed, rhythm, or sequence of actions, while subject variability reflects the diverse ways individuals perform the same activity due to physiology, behavior, or context. These factors make it difficult to build models that generalize well across users and situations. In unusual activity recognition for populations such as the developmentally disabled, these challenges are amplified by the short, unpredictable, and rare nature of unusual behaviors \cite{fujioka1,h}. 

To address these issues, deep learning architectures such as LSTMs, CNNs, and Transformer-based models are often used to capture temporal dependencies through sliding windows or sequence modeling. Leave-one-subject-out (LOSO) evaluation is commonly adopted to ensure generalization to unseen individuals. Approaches like convolution with self-attention (CSNet, TCCSNet) demonstrate that both local and global dependencies must be modeled, while self-supervised and transfer learning methods have been explored to improve robustness across diverse subjects and contexts \cite{e}.

\subsection{Language Models in Human Activity Recognition}

Large Language Models (LLMs) have opened new possibilities in HAR by addressing the scarcity of labeled data, particularly for rare or unusual activities that limit model training and generalization \cite{e}. Recent studies leverage LLMs to generate synthetic skeletal or inertial data and optimize parameters, enhancing robustness while preserving privacy \cite{f}. Techniques such as zero-shot and few-shot learning reduce annotation needs, while LLM-driven augmentation helps mitigate class imbalance and improve recognition in healthcare contexts \cite{f,g}. 

Despite these advantages, challenges remain. LLM-generated data may fail to capture inter-subject variability, introduce noise, or diverge from real distributions, while high computational cost and reliance on carefully designed prompts limit practicality \cite{e}. Future work is expected to combine LLM-based data generation with graph neural networks and sensor-specific transformers to better capture temporal and spatial dependencies \cite{h}. Advances in prompt design, multi-source integration, and cross-modal synthesis are also needed to improve data fidelity, reduce overhead, and ensure applicability in real-world monitoring and caregiving scenarios \cite{f,g}.

This paper presents the ISAS 2025 Challenge, which invites teams to address these limitations by adapting recent HAR solutions to unusual activity recognition for developmental disability support.

\section{Details of the Challenge}

\begin{figure}[h!]
    \centering
    \includegraphics[width=1\linewidth]{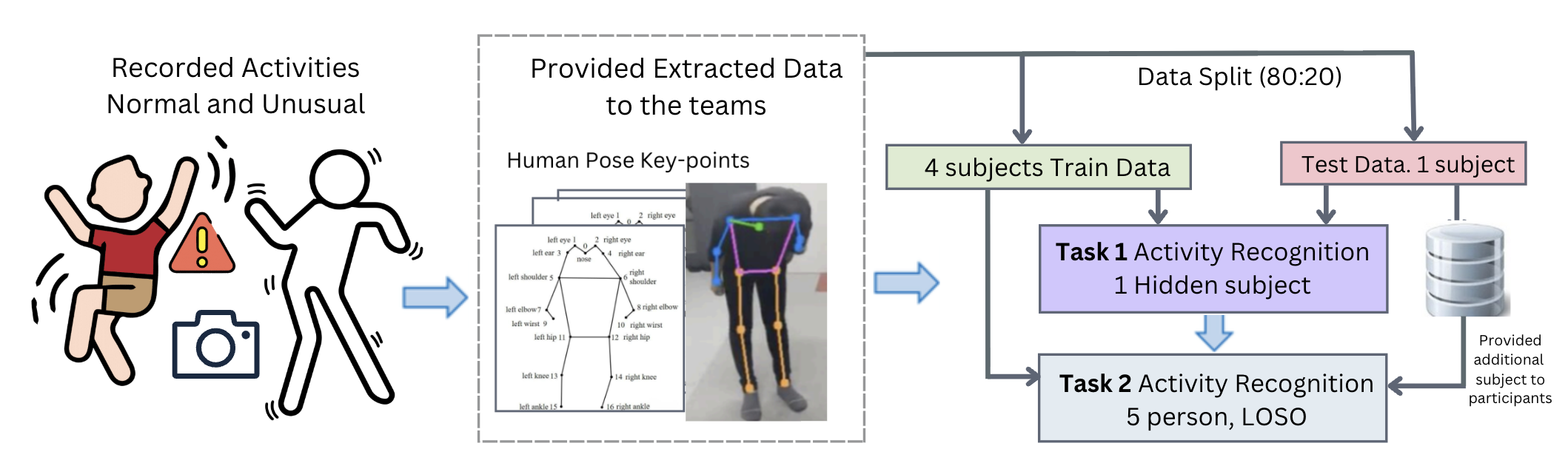}
    \caption{Challenge Overview  Dataset and with Two Tasks}
    \label{fig:challenge}
\end{figure}

The challenge focuses on recognizing unusual activities from pose-based data to support developmental disability care. Recorded normal and unusual  behaviors were converted into human pose keypoints and provided to participants as training and test datasets. Teams first classify activities of one hidden subject (Task 1) and then perform a five-subject Leave-One-Subject-Out (LOSO) evaluation (Task 2) to demonstrate model generalization as shown in Fig. \ref{fig:challenge}. The goal is robust multi-class activity recognition across eight defined activities, with performance evaluated primarily by Macro F1-Score to ensure fair assessment across all classes.

\subsection{Challenge Dataset}

The challenge dataset is based on 2D skeletal keypoints extracted from videos of healthy participants simulating eight activities relevant to developmental disability support. These include four \textit{normal} actions (eating snacks, sitting quietly, walking, and using a phone) and four \textit{unusual} behaviors (head banging, throwing objects, attacking others, and biting hands/fingers). The activities were designed in consultation with caregivers at a real care facility to ensure ecological validity and reflect behaviors that require timely monitoring and intervention.

For Task 1, two formats of the training dataset are shared to the participants:
\begin{itemize}
  \item \textbf{Option 1: Raw Keypoint Data with Timetable Log} \\
  Frame-by-frame 2D pose keypoints at 30 FPS with aligned timestamps and activity labels for four participants. This format is suitable for teams developing custom segmentation and temporal analysis techniques.

  \item \textbf{Option 2: Pre-Segmented and Labeled Samples} \\
  Labeled windows of pose keypoints for each activity sample, tagged as either \textit{normal} or \textit{unusual}, ideal for rapid prototyping and training.
\end{itemize}

A separate test dataset containing pose data from a fifth, unseen participant was initially provided for \textbf{Task~1} summarized in Table 1. Participants were required to classify each time step and submit a CSV file containing \texttt{participant\_id, timestamp, predicted\_label}. 

\begin{table}[h!]
\centering
\begin{tabular}{|l|l|l|}
\hline
\textbf{Activity} & \textbf{Behavior} & \textbf{Frames}\\
\hline
Sitting quietly & Normal & 24000 \\
Using phone & Normal & 23430 \\
Walking & Normal & 20190 \\
Eating snacks & Normal & 18810 \\
Head banging & Unusual & 5880 \\
Throwing things & Unusual & 3690 \\
Attacking & Unusual & 4350 \\
Biting nails & Unusual & 8490 \\
\hline
\end{tabular}
\caption{Activity Labels and Corresponding Behavior Type, Unseen 1 person Test Data}
\end{table}

Upon completion and submission of team predictions, the ground-truth labels of the unseen participant were released, allowing teams to proceed with \textbf{Task~2}. In this second phase, teams conducted a five-person Leave-One-Subject-Out (LOSO) evaluation, training on four participants and testing on the remaining one in rotation. From this evaluation, participants were expected to compute both per-person results and the average Accuracy and Macro F1-score across all folds, ensuring model generalization to unseen individuals.

\subsection{Limitations in the Data}

\begin{figure}[h!]
    \centering
    \includegraphics[width=0.50\linewidth]{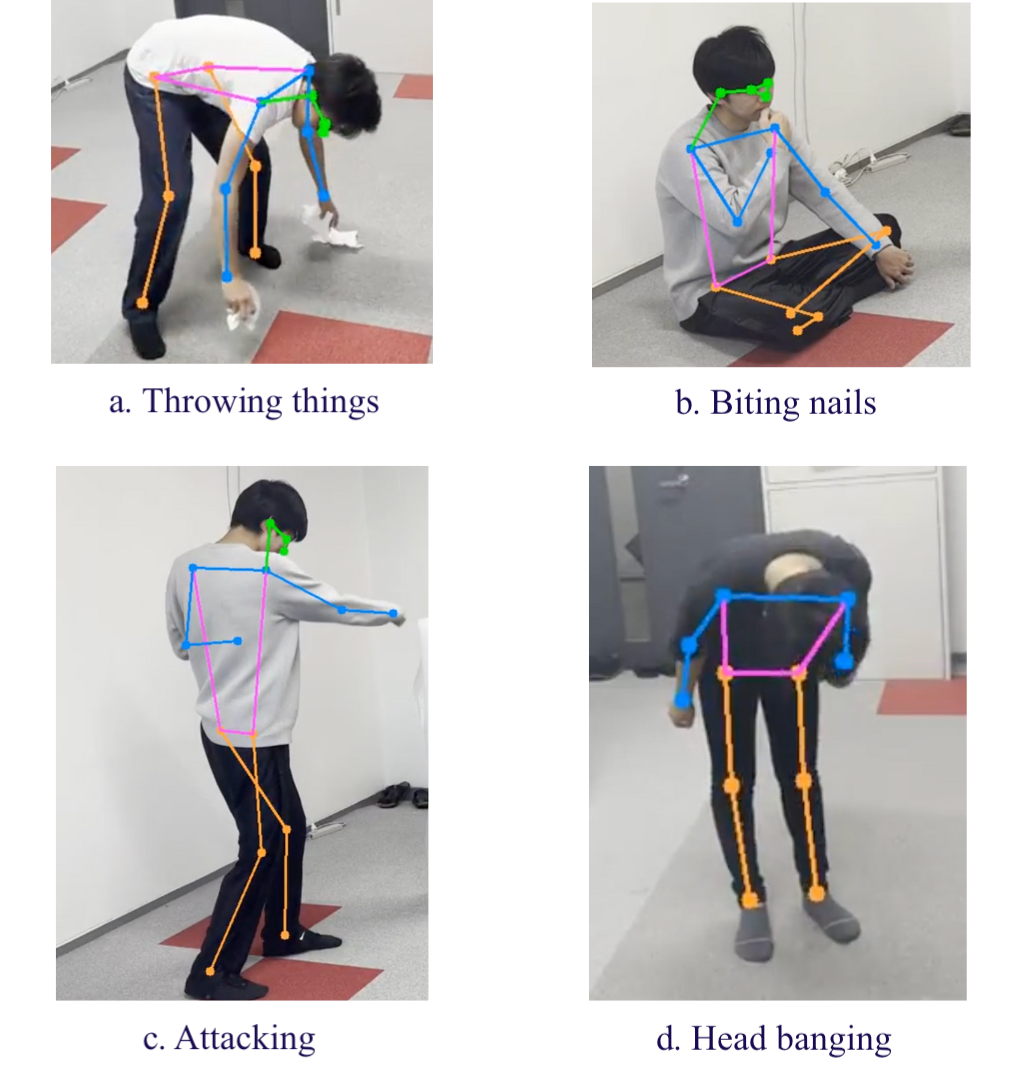}
    \caption{Examples of unusual behaviors represented in the dataset using pose estimation}
    \label{fig:Unusual}
\end{figure}

While the dataset provides useful insights into pose-based activity recognition for care facilities, several limitations should be acknowledged by teams. All actions were simulated by healthy participants, which may not fully capture the subtle motion patterns of individuals with developmental disabilities. Although these scenarios were designed in consultation with care staff, the ecological validity of the unusual behavior segments may be limited. Moreover, the dataset includes only five participants, restricting subject diversity and reducing the generalizability of trained models across broader populations. Examples of simulated unusual behaviors are illustrated in Fig.~\ref{fig:Unusual}.

Another important consideration is class imbalance; normal activities are significantly more frequent than abnormal ones. While this mimics real-world distributions, it also introduces challenges during training, especially for models sensitive to minority class performance. Unusual behaviors also tend to be abrupt and inconsistent in timing and form, making them harder to detect reliably through temporal analysis. Lastly, the dataset is limited to pose keypoints only—no visual, audio, or contextual environmental information is included. As a result, participants must rely solely on anonymized skeletal movement patterns, which may constrain the richness of behavior interpretation and classification accuracy.

\subsection{Goals of the Challenge}

This challenge focuses on developing models to recognize unusual behaviors from pose estimation data in a healthcare support context. Participants are tasked with distinguishing between normal and unusual activities using anonymized skeletal keypoint sequences derived from video recordings. The dataset includes eight activity types specifically four normal (sitting, eating, walking, using a phone) and four unusual (head banging, attacking, throwing objects, and biting fingers).

The primary objective is to build models that can accurately classify these activities on unseen participants’ data using a Leave-One-Subject-Out (LOSO) evaluation protocol. This simulates real-world deployment, where models must generalize to individuals not observed during training. The challenge reflects real-world complexities such as class imbalance (normal activities are more frequent), the abrupt and variable nature of unusual behaviors, and inter-subject differences in posture and activity expression.

The competition is open to all participants, with no restriction on team size. Supervisors may join multiple teams, but students affiliated with the host or organizing laboratories are not eligible. Multiple teams from the same laboratory must not privately share data or code, and in the case of team mergers, participants must inform the organizers directly.

Participants are encouraged to address the following dataset constraints into account when designing their models and interpreting performance outcomes, especially for real-world applications.

\begin{itemize}
  \item \textbf{Simulated Actors}. All activities were performed by healthy individuals imitating real behaviors, which may reduce realism and subtlety in motion.
  \item \textbf{Subject Diversity}. The dataset includes only five participants, which may limit the generalizability of models across wider populations.
  \item \textbf{Imbalanced Classes}. Normal behaviors dominate the dataset, mimicking real-world frequency but introducing challenges for training and evaluation.
  \item \textbf{Abruptness of Unusual Behaviors}. Unusual activities tend to start suddenly and have varying duration, increasing difficulty in detection.
  \item \textbf{Pose-Only Features}. No visual or audio data is provided, restricting analysis to anonymized skeletal sequences.
\end{itemize}

\subsection{Evaluation Protocol}

The evaluation is divided into two phases corresponding to Task 1 and Task 2. 

In \textbf{Task 1}, participants are required to submit predictions on a test dataset consisting of pose data from an unseen participant. Each time step must be classified into one of the eight predefined activity labels, and the results submitted in CSV format. Evaluation is based on standard classification metrics, including Accuracy, Precision, Recall, and F1-Score, with particular emphasis on the Macro F1-Score to ensure balanced performance across both normal and unusual activities.

For the test dataset, the goal is a multi-class classification. Participants must append an \texttt{"Action Label"} column to the test CSV file and populate it with one of the eight valid activity labels. Binary classification (e.g., just "Normal" vs "Unusual") is not acceptable.

\textbf{Submission Format:}
\begin{itemize}
  \item \textbf{Prediction CSV:} Must include \texttt{participant\_id}, \texttt{timestamp}, and \texttt{predicted\_label} (in a new column named \texttt{Action Label}).
  \item \textbf{Short Report:} A brief explanation of your approach and the LOSO evaluation findings.
  \item \textbf{Challenge Paper:} A complete paper (up to 12 pages including references) detailing methodology, LOSO results, visualizations (charts, confusion matrices, etc.), and key insights.
\end{itemize}

Upon completion of Task 1 and submission of predictions, the organizers release the ground-truth labels for the unseen participant along with the full dataset covering all five participants. In \textbf{Task 2}, teams must perform a Leave-One-Subject-Out (LOSO) evaluation, where models are trained on four participants and tested on the fifth in rotation. Teams are required to report per-participant results as well as the average Accuracy and Macro F1-Score across all folds. This two-phase setup ensures that models are evaluated both on their ability to classify activities for a single unseen individual and on their generalizability across multiple subjects.

During training and prediction, label consistency is crucial. Minor label variations such as \texttt{"Throwing"} and \texttt{"Throwing things"} must be treated as a single class. Additionally, training CSVs contain rows labeled \texttt{"None"} which refer to unrelated activities and must be excluded from the classification task. The model should focus only on the predefined eight activity classes.

At least one team member must present the paper during the workshop (hybrid). Accepted papers will be published in the ISAS proceedings. No private code or data sharing outside the registered team is allowed.

\section{Results of the Challenge}
Participants came from diverse academic and research institutions, showing growing international interest in behavior recognition using pose-based machine learning.

\subsection{Participating Teams}
\begin{figure}[h!]
    \centering
    \includegraphics[width=1.07\linewidth]{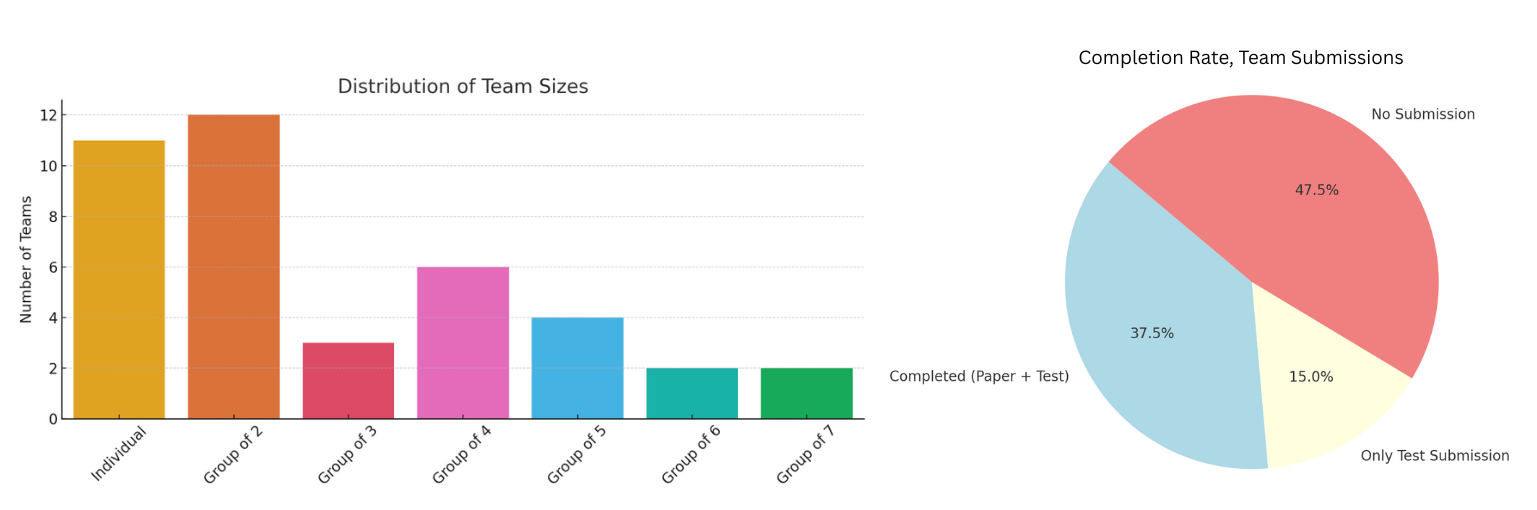}
    \caption{Team size distribution across participating groups (left) and challenge completion rate (right)}
    \label{fig:team_distribution}
\end{figure}

A total of 40 teams, comprising 97 participants from 11 countries, registered for the challenge. Most teams consist of 2 to 4 members, with some solo participants and one team reaching the maximum of 7 members, as shown in Fig.~\ref{fig:team_distribution}. Regarding challenge completion, 15 teams (37.5\%) successfully submitted both papers and experimental results, while 6 teams (15.0\%) provided only test predictions without corresponding papers. The remaining 19 teams (47.5\%) did not submit either predictions or papers, indicating varied levels of engagement across participants. This distribution reflects the challenge of balancing experimental work with paper preparation and highlights the selective nature of participation.

\begin{figure}[h!]
    \centering
    \includegraphics[width=0.65\linewidth]{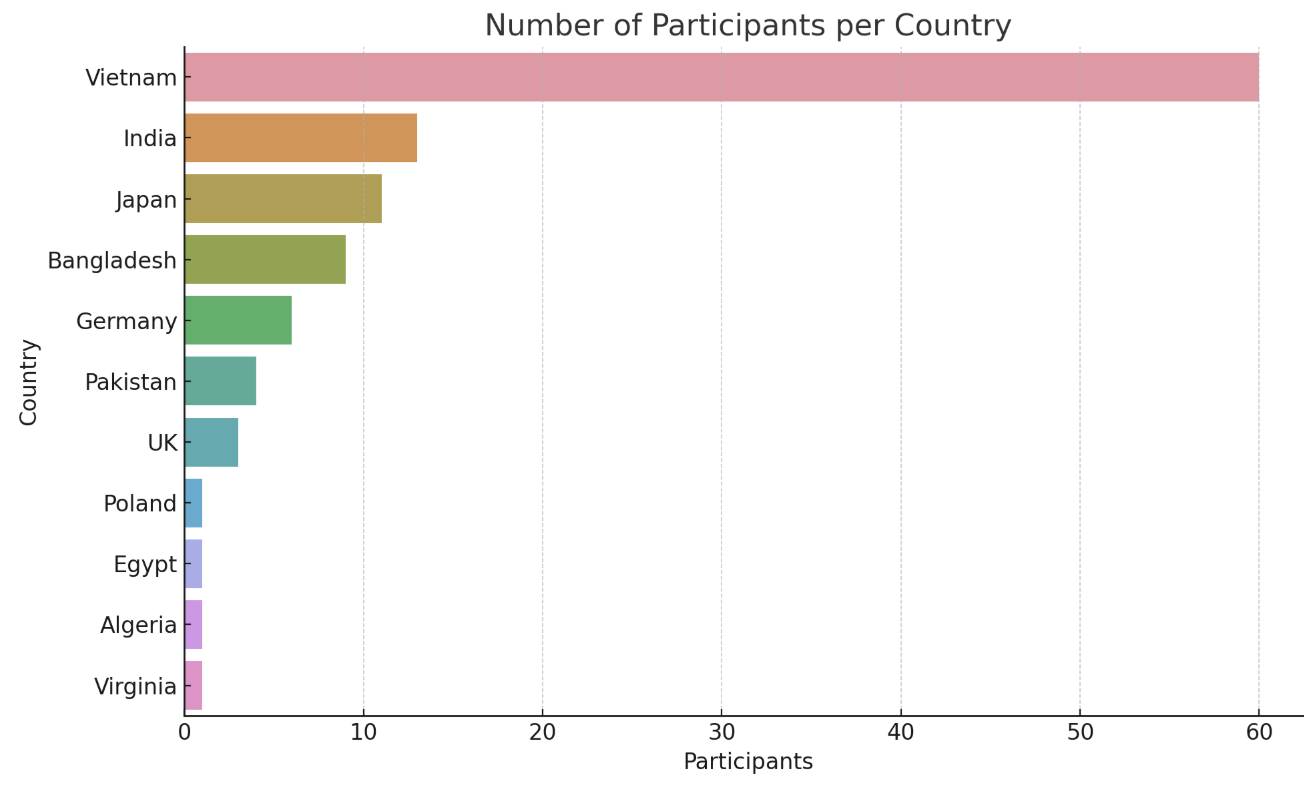}
    \caption{Country-wise distribution of participants. Vietnam had the highest number of participants, indicating strong regional engagement. Meanwhile, countries like Poland and the United Kingdom had the least number of participants, with only one representative each.}
    \label{fig:country_distribution}
\end{figure}

Across submissions, most teams focused on mitigating the dataset’s main challenges specifically class imbalance, temporal irregularity, and subject variability. Common strategies included overlapping sliding windows (typically 60–120 frames), ensemble learning, and weighted loss functions to boost sensitivity to rare abnormal events. Teams that combined statistical features with deep temporal models (CNN–LSTM, transformers, or adaptive ensembles) generally achieved higher F1-scores, while simpler pipelines relying only on classical ML without robust temporal modeling tended to underperform.

\subsection{Task 1: Predicting Unseen 1 Person Test Data}

To ensure fair comparison across teams, all submissions were evaluated on the same ground-truth frame set, with label variations normalized (e.g., \texttt{Throwing} merged with \texttt{Throwing things}). Frames labeled as \texttt{None} in the ground truth were excluded from scoring, while missing predictions were penalized. This standardized protocol allowed us to calculate accuracy, macro-precision, macro-recall, and macro-F1 consistently across all teams. The use of macro-averaging balanced the impact of class imbalance by giving equal weight to both frequent normal activities and rare unusual behaviors, ensuring that models sensitive to clinically significant abnormal activities were appropriately recognized.

As shown in Table 2, team performance in Task 1 varied widely, with \textit{Fake It Till We Make It} achieving the highest Macro F1-score (0.9198) and strong precision-recall balance, while several other teams demonstrated moderate performance in the 0.65–0.79 range. The lowest scores were observed for \textit{Frame\_Key} and \textit{Gundai Kiryu}, highlighting the difficulty of generalizing to unseen participants. These results emphasize both the potential and the challenges of unusual activity recognition from pose-based data.

\begin{table}[h!]
\centering
\begin{tabular}{|l|l|l|l|l|}
\hline
\textbf{Team Name} & \textbf{Accuracy} & \textbf{F1-Score} & \textbf{Precision} & \textbf{Recall}\\
\hline
Team Fake It Till We Make It & 0.8869 & \textbf{0.9198} & 0.9355 & 0.9261 \\
Team Bumblebee & 0.6996 & 0.7955 & 0.8436 & 0.7782 \\
Team Stormers & 0.7348 & 0.7752 & 0.8033 & 0.8095 \\
Team Binary Phoenix & 0.6811 & 0.7055 & 0.7418 & 0.7348 \\
Team hcmut\_lovers & 0.7501 & 0.7488 & 0.7729 & 0.7630 \\
Team TCL & 0.6701 & 0.6920 & 0.7837 & 0.6716 \\
Team TimelyCare & 0.6480 & 0.6824 & 0.6903 & 0.7434 \\
Team Du\_electrons & 0.6266 & 0.6540 & 0.6577 & 0.6612\\
Team waiedu\_isas& 0.6319 & 0.6525 & 0.6447 & 0.6873 \\
Team Nova Stellar & 0.6090 & 0.6463 & 0.6798 & 0.6524 \\
Team AI-MED Squad & 0.6083 & 0.5951 & 0.6817 & 0.5677 \\
Team ML-IOT & 0.5519 & 0.5481 & 0.5809 & 0.6203 \\
Team HCMUT D.I.C.T.S & 0.4403 & 0.3294 &0.3843 & 0.3793 \\
Team Gundai Kiryu & 0.1005 & 0.1038 & 0.185 & 0.1296 \\
Team Frame\_Key & 0.1810 & 0.0773 & 0.0761 & 0.1097 \\
\hline
\end{tabular}
\caption{Task 1 Performance of Teams on 1 Subject Test Data}
\end{table}

\begin{figure}[h!]
    \centering
    \includegraphics[width=1.05\linewidth]{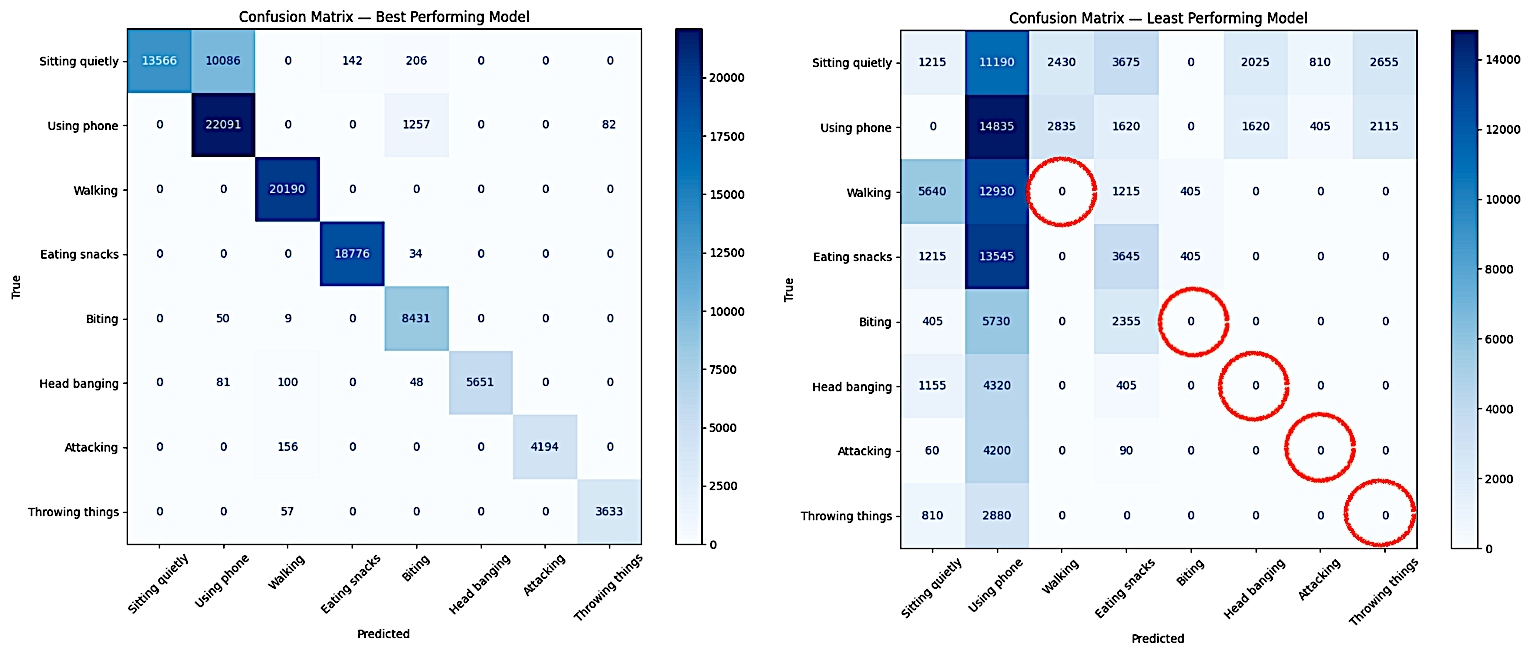}
    \caption{Task 1 Model Performance Comparison, One Unseen Person Test}
    \label{fig:cm_task1}
\end{figure}

The confusion matrix comparison in Fig. \ref{fig:cm_task1} highlights the contrast between the best and least performing models in classifying normal and unusual activities. With the best peforming model from Team Fake It Till You Make It, most activities are classified with very high accuracy, as indicated by the strong diagonal dominance. Even challenging unusual behaviors such as Attacking, Head banging, and Throwing things show substantial correct detection, though some minor misclassifications exist (e.g., Sitting quietly partially confused with Using phone).

The least-performing model (right confusion matrix in Fig. \ref{fig:cm_task1}  shows several limitations, particularly its failure to detect unusual behaviors such as biting, head banging, attacking, and throwing things, where almost all instances are misclassified into frequent normal classes like sitting quietly or using phone. Even normal activities such as walking and eating snacks are heavily confused with dominant classes, reflecting weak temporal modeling and poor class separation. These errors suggest overfitting to majority behaviors due to class imbalance and weak feature representation, leading the model to favor common patterns while neglecting rare but safety-critical activities which is an issue of particular concern in healthcare monitoring, where detecting unusual events with high sensitivity is essential.

\begin{figure}[h!]
    \centering
    \includegraphics[width=1.03\linewidth]{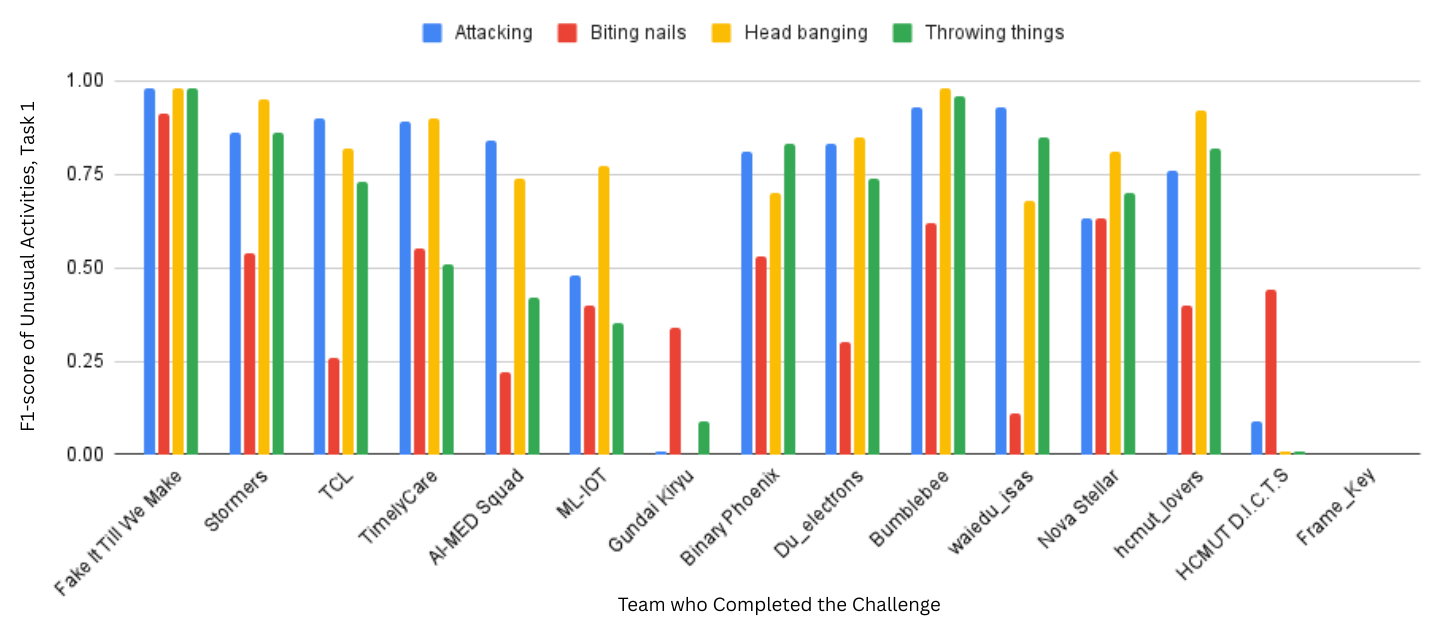}
    \caption{F1-scores for unusual activities, Task 1}
    \label{fig:F1_unusualT1}
\end{figure}

Investigating further, the results of the challenge show clear differences in how well teams were able to detect the unusual activities shown in Fig.~\ref{fig:F1_unusualT1}. Among the four classes, head banging was the most consistently recognized, with many teams achieving high F1-scores. This is likely because the motion is large, repetitive, and distinctive in the skeleton sequence, making it easier for models to identify. In contrast, biting nails was the least reliably detected, with very low F1-scores across most teams. This activity involves subtle hand-to-mouth gestures that are easily lost in pose estimation data and confused with normal activities such as sitting or using the phone. Attacking and throwing things showed moderate performance but with high variability between teams. These activities involve irregular and abrupt movements, which some models captured well, while others misclassified them as normal arm motions or walking. Overall, the figure illustrates that activities with clear, large-scale motion patterns are recognized more effectively, while fine-grained hand-based actions remain a major challenge, highlighting the need for models capable of capturing subtle temporal dynamics.


\subsection{Performance with 5 Participant Leave One Subject Out}

\begin{figure}[h!]
    \centering
    \includegraphics[width=1.00\linewidth]{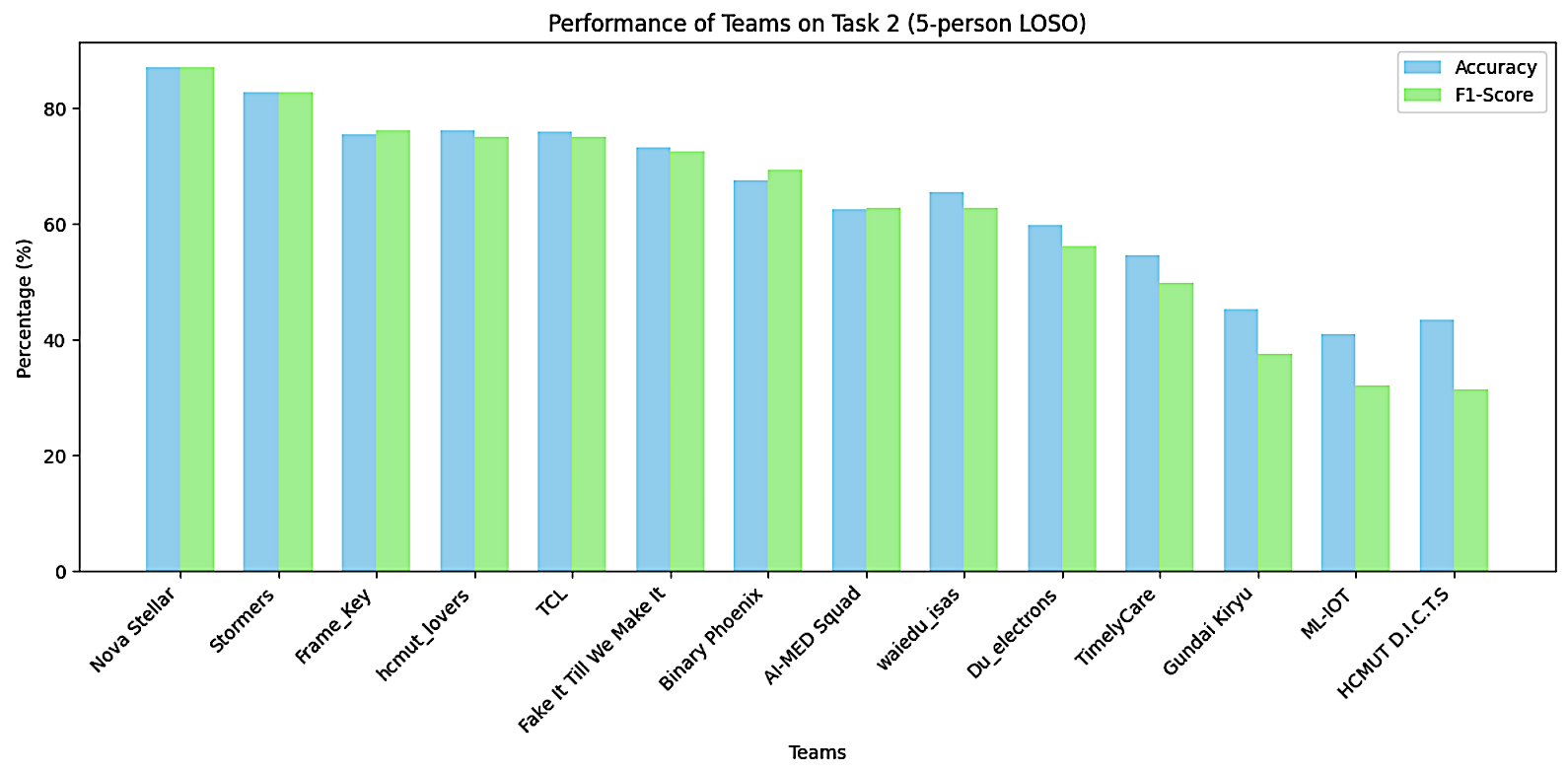}
    \caption{Average F1-score and Accuracy by Teams on 5-person LOSO, Task 2}
    \label{fig:Task2}
\end{figure}

The LOSO evaluation results is summarized in Fig. \ref{fig:Task2} with a wide range of team performances. Team Nova Stellar achieved the highest accuracy and F1-score (both at 87\%), demonstrating strong model generalization across unseen subjects. Stormers also performed well (Acc: 82.8\%, F1: 82.7\%), suggesting their deep learning approach effectively captured temporal dynamics.

A consistent pattern across submissions is that F1-scores are slightly lower than accuracy, reflecting challenges in recognizing rare abnormal behaviors compared to dominant normal ones. Teams with ensemble or deep temporal models tend to outperform purely statistical or basic ML methods, reinforcing the importance of robust temporal modeling and data representation for this task.

\section{Summary of the Implemented Methodologies}

From an implementation standpoint, Python (v3.8–3.10) was the standard language, with PyTorch, TensorFlow, and scikit-learn as the dominant libraries. Most teams trained on Google Colab with Tesla T4/P100 GPUs, while others used local GPUs such as RTX 30-series or cloud CPUs. The strongest performance came from Fake It Till We Make It, which achieved the best F1-score through an ML+DL ensemble with 60-frame segmentation. On the other hand, Frame\_Key and Gundai Kiryu showed the weakest results, largely due to minimal temporal modeling and reliance on basic classifiers. Overall, these results highlight that ensemble and deep learning approaches, especially those that take advantage of temporal context, are more effective in achieving robust detection of unusual activities in this benchmark.

\begin{table}[htbp]
\centering
\caption{Team methods and settings for Task 1 (Unseen Subject)}
\vspace{0.8em}
\begin{tabular}{|p{2.2cm}|p{2cm}|p{4.5cm}|p{2.3cm}|p{3cm}|}
\hline
\textbf{Team} & \textbf{Method (ML/DL)} & \textbf{Short Detail Method} & \textbf{Window Size} & \textbf{Hardware} \\
\hline
Stormers & Deep Learning & A deep learning model combining stacked temporal modules (with segmentation) & 120 frames (75\% overlap) & RAM: 32 GB, CPU: Intel CORE i7 @ 2.40 GHz, GPU: GeForce RTX 4060 \\
\hline
TCL & Machine Learning & HistGradientBoostingClassifier with a rich hand-crafted + auto time-series feature set & $\sim$5 s windows ($\sim$150 frames), overlapping & RAM: 12 GB, CPU: AMD Ryzen 7 7000 Series, GPU: NA \\
\hline
TimelyCare & Statistical + Deep Learning & Statistical features feeding a lightweight DL head (with dropout) & 90 frames & 4 GB RTX 3050 \\
\hline
Fake It Till We Make It & ML + DL Ensemble & Ensemble with stride-based segmentation over 60-frame sequences & 60 frames & RAM: 32 GB, CPU: AMD Ryzen 5 5600X, GPU: RTX 3070 / 3060 \\
\hline
Gundai Kiryu & ML + DL (basic) & Simple sequence splitting and baseline classifiers & Variable & RAM: 16 GB, CPU: Intel CORE i7 @ 2.4 GHz, GPU: NA \\
\hline
ML-IOT & Statistical + LSTM & Statistical features with LSTM models, 50\% overlap & 30 frames (50\% overlap) & Google Colab (T4) \\
\hline
AI-MED Squad & Statistical + ML (feature selection) & Hand-crafted features with classical ML and feature selection & $\sim$5 s windows & CPU: 11th-Gen Intel i7-1165G7, 16 GB RAM \\
\hline
Binary Phoenix & CNN–LSTM Hybrid & CNN–LSTM pipeline with data augmentation on 60-frame clips & 60 frames & Apple M1 Pro (16 GB Unified) \\
\hline
Du\_electrons & ML/DL Hybrid & Overlap segmentation with hybrid learner & 90 frames (50\% overlap) & CPU: AMD Ryzen 5 5600H; GPU: RTX 3050 Ti \\
\hline
Team \newline Bumblebee & Adaptive Ensemble & Adaptive windowing with statistical + ML/DL ensemble & Variable/ adaptive & Intel Xeon, 31 GB RAM (cloud) \\
\hline
waiedu\_isas & Statistical + ML & Statistical descriptors with classical ML; temporal overlap & (noted in text) & Google Colab \\
\hline
Nova Stellar & Deep Learning & Hybrid transformer-style temporal network on pose sequences & (noted in text) & Google Colab (Tesla T4/P100) \\
\hline
hcmut\_lovers & Machine Learning & CatBoost-based classifier with engineered temporal features & (noted in text) & RAM: 13 GB, CPU: Intel Xeon @ 2.20 GHz, GPU: Tesla T4 \\
\hline
HCMUT D.I.C.T.S & ML + DL (mixed) & Combined classical ML and DL baselines & (noted in text) & RAM: 8 GB, CPU: Intel CORE i5 @ 2.4 GHz, GPU: NA \\
\hline
Frame\_Key & ML + DL (mixed) & Cascaded classical models (e.g., XGB) with sliding window & 15 s (overlap $\approx$ 1.5 s) & Google Colab \\
\hline
\end{tabular}
\label{tab:task1_methods}
\end{table}

As shown in Table~\ref{tab:task1_methods}, most participating teams adopted hybrid approaches that combined 
\textbf{machine learning (ML)} with \textbf{deep learning (DL)} or statistical features, reflecting a balance between 
handcrafted feature engineering and modern sequence modeling. The most common strategy was the use of 
overlapping sliding windows of 60--120 frames, often paired with ensemble methods or CNN--LSTM architectures. 
Among all submissions, the team \textit{Fake It Till We Make It} achieved the \textbf{highest performance} 
(Macro F1 = 0.9198) using a ML + DL ensemble with stride-based segmentation on 60-frame windows, supported 
by high-end GPUs (RTX 3070/3060). In contrast, the weakest-performing teams, such as \textit{Frame\_Key} 
(Macro F1 = 0.0773) and \textit{Gundai Kiryu} (Macro F1 = 0.1038), relied on basic ML/DL mixes or simplistic 
windowing strategies, often with minimal GPU support. The results also indicate that \textbf{hardware resources 
influenced performance} with teams running on dedicated GPUs or cloud accelerators (e.g., RTX 4060, Tesla T4, 
RTX 3070) generally outperformed those constrained to CPUs or limited RAM. Notably, statistical + ML pipelines 
(e.g., \textit{AI-MED Squad}, \textit{waiedu\_isas}) achieved moderate results (Macro F1 $\approx$ 0.59--0.65), 
showing that while classical methods remain competitive, the \textbf{top-performing systems leveraged deeper 
temporal modeling and ensemble designs}.

Team Fake It Till You Make It demonstrated good performance strongest performance, achieving the good Macro F1-score in both Task 1 and Task 2. Their approach stood out by combining two complementary strategies: a lightweight TinyHAR pose-based model with sliding window configurations, and a transformation of pose data back into video sequences for testing video classification models. This dual strategy allowed for both efficiency and robustness. Importantly, their analysis extended beyond accuracy, incorporating comparisons of GPU VRAM and power consumption, which highlighted the practicality of TinyHAR for deployment on resource-limited devices such as low-cost laptops. Compared with video-based baselines like R(2+1)D, TinyHAR offered superior energy efficiency while maintaining high accuracy, showing strong potential for real-world applications.

Overall, the challenge also revealed several methodological lessons. First, there was no consensus on optimal window size, with teams using segments from 30 to over 120 frames implying shorter windows better captured abrupt unusual events, while longer ones preserved context. Second, classical ML with feature engineering and deep learning pipelines each showed distinct strengths, suggesting that hybrid designs may be most effective. Third, explicit handling of class imbalance, such as weighted losses, proved critical for detecting rare behaviors. Finally, the LOSO evaluation confirmed that rigorous subject-agnostic validation is essential to avoid overfitting to individual motion styles.

\section{Conclusion}
The ISAS 2025 Unusual Activity Recognition Challenge
for Developmental Disability Support demonstrated both the promise and difficulty of pose-based unusual behavior recognition in care contexts. While teams explored methods from classical ML to CNN–LSTM hybrids and transformer-based models, results showed that class imbalance, abrupt activity transitions, and subject variability remain major hurdles. The top-performing teams combined TinyHAR with sliding windows, video-classification pipelines, and class weighting to achieve the best Macro F1-scores, while also proving efficiency for real-world deployment on limited hardware. Overall, the challenge not only benchmarked effective strategies but also underscored the need for models that generalize robustly across individuals and balance performance with deployability, offering key insights for advancing socially responsible AI in healthcare monitoring.

\end{document}